\documentclass[letterpaper, 10 pt, conference]{ieeeconf}
\usepackage{amsmath,amsfonts}
\usepackage{array}
\usepackage{textcomp}
\usepackage{stfloats}
\usepackage{url}
\usepackage{verbatim}
\usepackage{graphicx}
\usepackage{cite}
\usepackage{xcolor}
\usepackage{subcaption}
\usepackage{mathtools}
\usepackage{amssymb}
\usepackage{tabulary}
\usepackage{booktabs}
\usepackage[ruled,linesnumbered]{algorithm2e}
\usepackage{setspace}
\usepackage{multirow}
\usepackage{enumerate}
\makeatletter
\let\NAT@parse\undefined
\makeatother
\usepackage{hyperref}
\usepackage{makecell}
\usepackage{cuted}
\usepackage{capt-of}

\IEEEoverridecommandlockouts
\title{\LARGE \bf
SparseGrasp: Robotic Grasping via 3D Semantic Gaussian Splatting from Sparse Multi-View RGB Images
}

\author{Junqiu Yu$^{1*}$, Xinlin Ren$^{1*}$, Yongchong Gu$^{1}$, Haitao Lin$^{1}$, Tianyu Wang$^{1}$, Yi Zhu$^{2^\dagger}$,\\ Hang Xu$^{2}$, Yu-Gang Jiang$^{1}$, Xiangyang Xue$^{1}$, Yanwei Fu$^{1^\dagger}$ 
\thanks{$^\dagger$: Corresponding Authour.}
\thanks{$*$: Equal Contribution.}
\thanks{$^{1}$ Junqiu Yu, Xinlin Ren, Yongchong Gu, Haitao Lin, Tianyu Wang, Yu-Gang Jiang, Xiangyang Xue and Yanwei Fu are with Fudan University. { \{jqyu20, xlren20,htlin19,ygj ,xyxue,yanweifu\}@fudan.edu.cn}, { \{yongchonggu22, tywang22\}@m.fudan.edu.cn}}%
\thanks{$^{2}$ Yi Zhu and Hang Xu are with the Department of Noah's Ark Lab, Huawei Technology, Shanghai 200433, China. { \{zhuyi36, xu.hang\}@huawei.com}}%
}


\begin{document}
\maketitle
\begin{strip}\centering
\vspace{-0.4in}
\includegraphics[width=\textwidth]{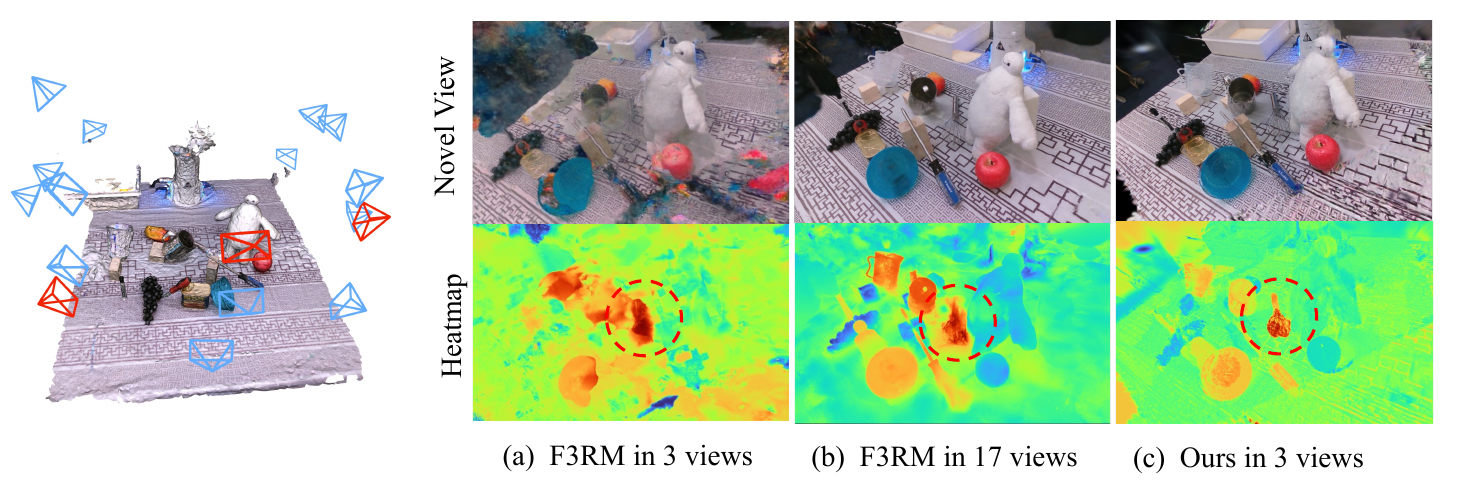}
\captionsetup{font={small}}
\captionof{figure}{We present a comparison between our SparseGrasp and F3RM under both sparse and dense view settings. The top row shows the novel view images, while the bottom row displays the heat map of feature field using the text ``whisk" as a query. Remarkably, our method, utilizing only \textbf{3} view images, achieves performance comparable to F3RM, which is trained with 17 views.}
  \label{fig:teaser}
\end{strip}

\begin{abstract}

Language-guided robotic grasping is a rapidly advancing field where robots are instructed using human language to grasp specific objects. However, existing methods often depend on dense camera views~\cite{shen2023Distilled, lerftogo2023} and struggle to quickly update scenes, limiting their effectiveness in changeable environments.
In contrast, we propose SparseGrasp, a novel open-vocabulary robotic grasping system that operates efficiently with sparse-view RGB images and handles scene updates fastly.
Our system builds upon and significantly enhances existing computer vision modules in robotic learning. Specifically, SparseGrasp utilizes DUSt3R~\cite{wang2024dust3r} to generate a dense point cloud as the initialization for 3D Gaussian Splatting (3DGS), maintaining high fidelity even under sparse supervision. Importantly, SparseGrasp incorporates semantic awareness from recent vision foundation models. To further improve processing efficiency, we repurpose Principal Component Analysis (PCA) to compress features from 2D models. Additionally, we introduce a novel render-and-compare strategy that ensures rapid scene updates, enabling multi-turn grasping in  changeable environments.
Experimental results show that SparseGrasp significantly outperforms state-of-the-art methods in terms of both speed and adaptability, providing a robust solution for multi-turn grasping in changeable environment. 


\end{abstract}


\section{Introduction}



Language-guided robotic grasping is an emerging field where robots use human language instructions to grasp specific objects. Imagine a scenario where a robot can swiftly and reliably grasp objects based on verbal commands while simultaneously adapting to changes in the environment. This would enable the robot to seamlessly follow up on new instructions. Achieving this requires the robot to not only locate objects accurately through language but also to understand their geometry, ensuring more precise and effective grasping.

Recent efforts~\cite{shen2023Distilled, lerftogo2023, li2024object, ze2023gnfactor} aim to reconstruct scenes while distilling semantic information from 2D foundation models like CLIP~\cite{radford2021learning}. However, these methods heavily rely on dense view reconstruction, which demands significant training time and multi-view capturing via the robot’s on-board camera. Some viewpoints are challenging for robotic arms to reach, requiring precise path planning, while performance drops when only limited views are available. For instance, as shown in Fig.~\ref{fig:teaser}, F3RM~\cite{shen2023Distilled} struggles with object geometry using only 3 views, and even with 17 views, semantic distillation remains suboptimal. Additionally, these methods are designed for static environments, requiring dense view capture even for minor changes. This hinders their ability to adapt to changeable environments and perform multi-turn grasping, limiting their real-world applicability. Here, `changeable environment' refers to scenarios where objects may be moved, while `multi-turn' indicates the robot executing a sequence of language commands.

 To address these challenges, we introduce SparseGrasp, which enables fast scene reconstruction (around 240 seconds) and fast updation using only \textbf{sparse-view RGB images}. Since 3D Gaussian Splatting (3DGS)~\cite{kerbl20233d} tends to overfit with sparse views, we integrate DUSt3R~\cite{wang2024dust3r} to generate a dense point cloud as initialization, offering greater robustness than sparse point clouds, as  in Fig.~\ref{fig:dense_point}. 
Besides, we further incorporate 2D foundation models like MaskCLIP~\cite{zhou2022extract} and Segment Anything (SAM)~\cite{kirillov2023segment} to extract dense semantic features, which are distilled into 3DGS. Unlike prior methods~\cite{lerftogo2023} that require multiple CLIP calls for each object identified by SAM, we propose a more efficient approach: applying patch-level CLIP and SAM once per image to generate comprehensive semantic features.
Further, given the high dimensionality of semantic features, directly distilling them into 3DGS is impractical. Therefore, we use PCA to compress the features, reducing dimensions significantly (from 768 to 16). 
In addition, we improve GraspNet~\cite{fang2020graspnet} by generating grasps directly from 3DGS, eliminating the voxelization and depth back-projection required by methods like F3RM~\cite{shen2023Distilled} and LERF-TOGO~\cite{lerftogo2023}. Finally, for objects that have changed positions in the scene, we propose a render-and-compare strategy to efficiently update their scene representations, eliminating the need for full scene reconstruction.

Our contributions are as follows: 1) We present the \textit{SparseGrasp} system for rapid scene reconstruction (240 seconds) and fast updates using sparse-view RGB images, overcoming the dense-view dependency in prior methods. 2) We propose \textit{3D Semantic Gaussian Splatting}, incorporating DUSt3R for robust dense point cloud initialization, and enhancing semantic distillation by efficiently applying MaskCLIP and SAM once per image to extract dense semantic features, followed by PCA for feature reduction. 3) We improve GraspNet by generating grasps directly from 3DGS, eliminating voxelization and depth back-projection. 4) Finally, we introduce a \textit{render-and-compare strategy} for efficiently updating scene representations when objects change, avoiding the need for full scene reconstruction.

\section{Related Work}

\noindent{\textbf{Rendering Methods for Robotics.}}
Neural Radiance Fields (NeRF)\cite{mildenhall2021nerf} have recently advanced to enable high-quality scene reconstruction from RGB images, leading to their integration into various robotics applications such as grasping\cite{kerr2023evo, ichnowski2021dex, dai2023graspnerf, yen2022nerf} and navigation~\cite{sucar2021imap, zhu2022nice, rosinol2023nerf, adamkiewicz2022vision}. Despite their exceptional reconstruction quality, NeRF methods typically focus on complete scene reconstruction and are limited by the time-consuming process of capturing multiple images and training models, making them suitable only for static environments. While Evo-NeRF~\cite{kerr2023evo} aimed to speed up scene updates for sequential grasping, it compromised on object geometry accuracy and still required multiple perspectives, reducing its efficiency for language-guided grasping.
In contrast, 3DGS~\cite{kerbl20233d} offers efficient scene reconstruction through adaptive density control and multi-view images, proving valuable in SLAM~\cite{keetha2023splatam, yugay2023gaussian, yan2023gs}. However, its application to robotic grasping has been limited till now, with notable use only in simulated environments~\cite{lu2024manigaussian}. 
%
Our approach innovatively applies 3DGS to \textit{RGB images from sparse camera views}, achieving faster and more efficient scene reconstruction and updates. This enables precise grasping of objects in both static and dynamically changing environments. To the best of our knowledge, SparseGrasp is the first method to leverage only \textit{sparse view RGB images} for scene reconstruction and object grounding in changeable environment.

\noindent{\textbf{Language-guided Grasping.}}
The integration of Computer Vision (CV) and Natural Language Processing (NLP) enhances robots' language comprehension. Early studies~\cite{guadarrama2014open, hatori2018interactively, shridhar2018interactive, nguyen2020robot, chen2021joint} achieved object grounding with 2D models. More recent approaches~\cite{cheang2022learning, sun2023language} combine visual grounding with 6D pose estimation~\cite{lin2022sar} for robotic grasping, though they may struggle with precision in fine-grained object manipulation.
%

%

Recent advancements have integrated semantic information into 3D representations. Works like CLIPort~\cite{shridhar2022cliport} align semantic features with point clouds or scene depth, but rely exclusively on depth data, resulting in suboptimal alignment. Other approaches~\cite{shen2023Distilled, lerftogo2023, ze2023gnfactor} improve semantic information embedding within 3D scene reconstruction. Specifically, F3RM~\cite{shen2023Distilled} and LERF-TOGO~\cite{lerftogo2023} train NeRF using RGB images, while GNFactor~\cite{ze2023gnfactor} leverages RGB-D data to generate voxelized scene representations and aligns semantic features with voxels. However, these methods may struggle when scene changes and require additional time to adapt.
Our method uniquely leverages 3DGS to learn semantic information directly from sparse RGB images and employs render-and-compare strategy, enabling robots to perform sequential operations effectively even in changeable environment.


\section{Method}
\begin{figure*}
\begin{center}
\includegraphics[width=\linewidth]{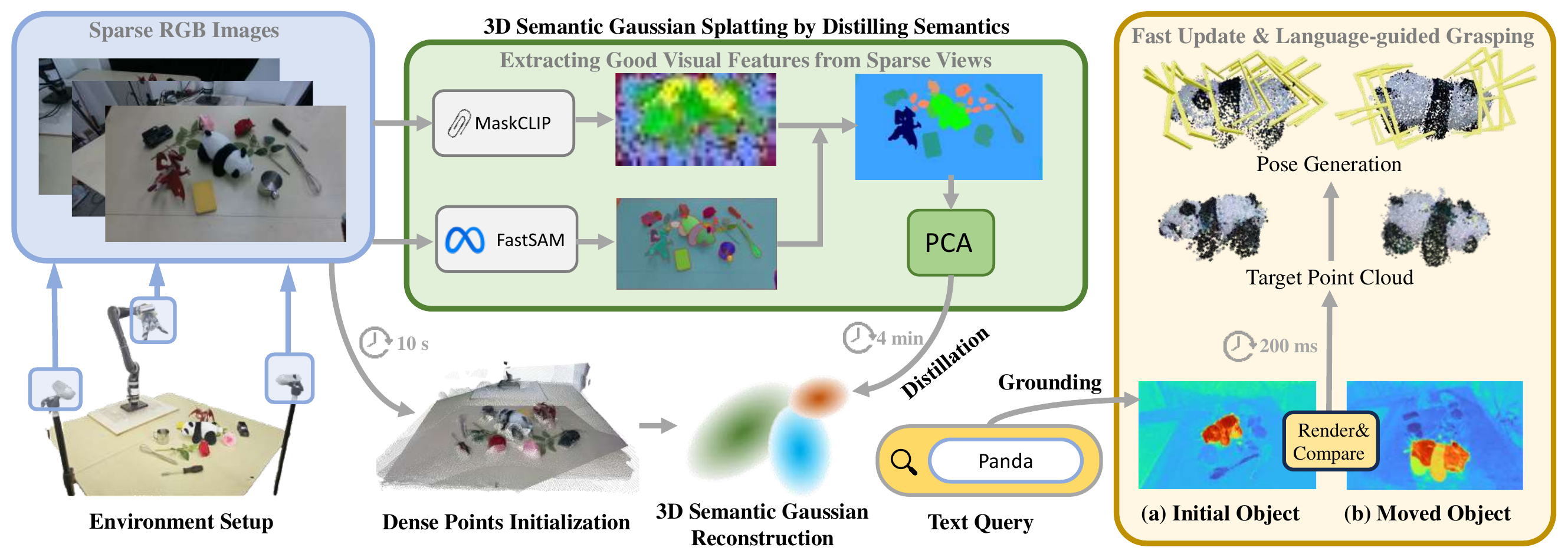}
\end{center}
\vspace{-0.1in}
\caption{Our architecture. It starts with collecting sparse view images and generating dense point clouds to initialize 3DGS. Next, we integrate FastSAM and MaskCLIP to generate average features within each mask. Then,  PCA is applied to compress the whole average features in a low dimension,  then distilled into 3DGS. Given an open-vocabulary language instruction, our system can locate the target object and generate appropriate grasp poses. When scene changes, the Render-and-Compare strategy enables fast scene updates. \label{fig:pipeline} }

\end{figure*}

Given sparse view images $I$, our goal is to continuously pick and place objects according to the open-vocabulary languages in static and changeable scenes. Formally, this involves generating 6-DoF grasp poses $\mathbf{T} = (R, t)$ where $R,t$ represent the rotation and translation components of the grasp pose, respectively. 
%
%
Our approach interprets language commands to guide robotic actions in diverse and changing environments. In Fig.~\ref{fig:pipeline}, we start by collecting RGB images from sparse viewpoints and use DUSt3R~\cite{wang2024dust3r} to initialize 3DGS with dense point clouds. We then integrate semantic features into 3DGS, focusing on the extraction and compression of pixel-level CLIP features, and employ the Render-and-Compare strategy for rapid scene updates. Additionally, we introduce language-guided robotic grasping.




\noindent \textbf{Preliminary: 3D Gaussian Splatting.}
3DGS reconstructs scenes from images by approximating objects in 3D space using Gaussians, initialized with point clouds generated by COLMAP~\cite{schonberger2016structure} from dense view images.
Each 3D Gaussian $g_i$ is characterized by a 3D coordinate $p_i \in \mathbb{R}^3$, a scaling factor $s_i \in \mathbb{R}^3$, a rotation quaternion $q_i \in \mathbb{R}^4$, and an opacity value $\alpha_i \in \mathbb{R}$. Color $c_i \in \mathbb{R}^3$ can be derived from spherical harmonics with given direction. Using tile-based rasterization, the pixel color $C$ is calculated:
\begin{align}
    C &= \sum_{i\in \mathcal{N}} T_i c_i \alpha_i 
\end{align}
\noindent where $T_i = \prod_{j=1}^{i-1}(1-\alpha_j)$. 
During the training phase, the loss is computed to optimize the parameters. Leveraging this rasterization technique, 3DGS facilitates real-time rendering.



\subsection{\textbf{Extracting Good Visual Features from Sparse Views}}

\begin{figure*}[htbp]
\begin{center}
\includegraphics[width=\linewidth]{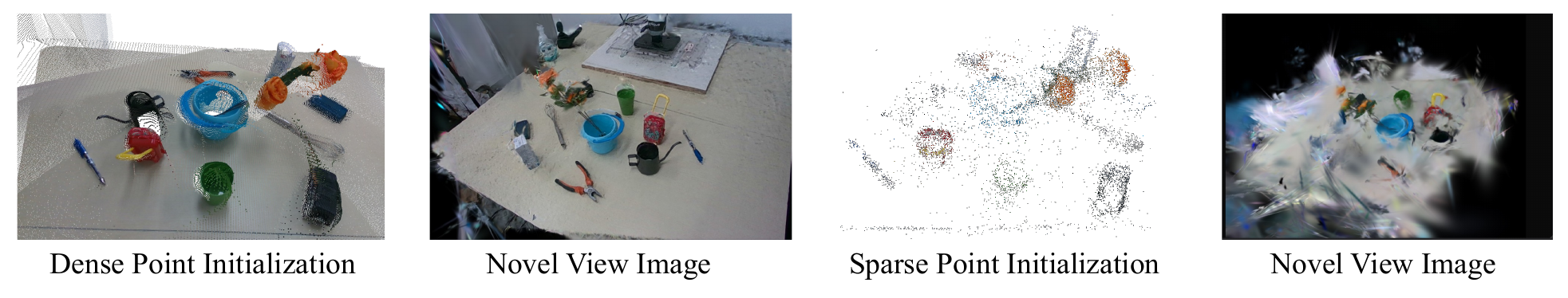}
\end{center}
\vspace{-0.1in}
\caption{Comparison of dense point initialization v.s. sparse point initialization in sparse view images. Initializing with sparse points often leads to overfitting with sparse view images. }\label{fig:dense_point}
\end{figure*}


\noindent \textbf{Dense Point Initialization}. 
We initialize with dense points using DUSt3R~\cite{wang2024dust3r} to generate dense point clouds
in sparse view scene reconstruction.
Traditional 3DGS requires dense-view images and uses COLMAP~\cite{schonberger2016structure} to create sparse point clouds, which can be problematic in sparse-view settings. NeRF-based methods also tend to overfit with sparse-view images. While many methods \cite{zollhofer2018state, wang2014robust} use RGB-D sensors to supplement depth information, this adds hardware requirements. Our approach overcomes these limitations by enabling fast reconstruction from sparse views and demonstrates robustness in novel views, as illustrated in Fig.~\ref{fig:dense_point}.

\noindent{\textbf{Extracting Dense Visual Features.}}
To extract dense visual features, we address two key challenges: 1) achieving per-pixel feature extraction with clear object boundaries, and 2) minimizing time expenditure. 

To overcome CLIP's limitation of extracting features only at the image level, we adapt the approach from F3RM~\cite{shen2023Distilled} by using the MaskCLIP~\cite{zhou2022extract} reparameterization trick for patch-level alignment. However, patch-level CLIP features alone often yield imprecise object boundaries, complicating accurate object grounding.

So, instead of using DINO features for regularization~\cite{kerr2023lerf} or invoking CLIP on numerous masked segments from SAM~\cite{kirillov2023segment}, which can be inefficient, our method extracts patch-level features with MaskCLIP and applies nearest neighbor upsampling to match the input image resolution. We then average the features within each mask generated by FastSAM~\cite{zhao2023fast}, producing a dense feature matrix $\mathbf{F}^{\mathrm{sem}} \in \mathbb{R}^{N \times C}$, where N is the number of masks.

A key consideration is that the masks produced by FastSAM may overlap. To resolve this, our method first sorts the masks by size in ascending order, ensuring that smaller masks are prioritized. We then create a new mask where the smaller masks take precedence in overlapping regions by selecting the first non-zero mask at each pixel. It ensures that the final mask resolves any overlaps by favoring smaller regions. Through this approach, we successfully extract precise and dense pixel-level semantic features at around 180ms. In addition, since our method requires only sparse views as input, the time spent on feature extraction is negligible.

\subsection{\textbf{3D Semantic Gaussian Splatting by Distilling Semantics \label{sec:sparse_rec}}}
\noindent{\textbf{Compression of Language Features.}}
Unlike NeRF, 3DGS uses over 100,000 Gaussians, leading to high memory and computation costs with high-dimensional CLIP features. While Feature-3DGS~\cite{zhou2023feature} uses a lightweight decoder and LangSplat~\cite{qin2023langsplat} employs a scene-specific autoencoder, both methods either reduce rendering speed or require extensive training (over 30 minutes).
In contrast, our method uses PCA for efficient feature compression. By averaging object features, we apply PCA to reduce the feature set to as few as 16 dimensions, maintaining sufficient accuracy for scene representation, as illustrated in Fig.~\ref{fig:pca}. The compressed feature is denoted as $\mathbf{F}^{\mathrm{sem}}_{pca}$.

\noindent{\textbf{Distilling Semantic Features into 3D Gaussians.}} 
%
%
We enhance 3DGS by integrating language features into each 3D Gaussian. Specifically, we use differential rasterization to derive dense semantic features $\mathbf{\hat{F}}^{sem}$ for each pixel.
\begin{align} 
    \hat{F}_{i}^{sem} = \sum_{i\in N}f_i\alpha_i T_i
\end{align}
%
Where, $f_i$ denotes the semantic feature within each 3D Gaussian. Rather than rasterizing the RGB image and feature map separately, we use a joint optimization approach where both are processed through the same tile-based rasterization. Unlike LangSplat~\cite{qin2023langsplat}, which calculates the gradient of $\alpha$ using language features, our method derives this gradient solely from RGB images, given their more reliable supervision. We optimize our 3DGS with the following objective:
\begin{align}
    \mathcal{L} &= \mathcal{L}_{rec} + \lambda_2 \cdot \mathcal{L}_{sem}, \\
    \mathcal{L}_{rec} &= (1 - \lambda_1)\mathcal{L}_1 + \lambda_1\cdot\mathcal{L}_{\mathrm{D-SSIM}}, \\
     \mathcal{L}_{sem} &= \mathcal{L}_1(\mathbf{F}^{\mathrm{sem}}_{pca}, \hat{\mathbf{F}}^{\mathrm{sem}})
\end{align}
where $\mathcal{L}_1$ denotes the L1 loss and $\mathcal{L}_{\mathrm{D-SSIM}}$ is the SSIM loss between the rendered image and the ground truth. The reconstruction loss $L_{rec}$ is formulated as in the original 3DGS~\cite{kerbl20233d}, and $L_{sem}$ represents the L1 loss between the rendered and compressed semantic features. The terms $\lambda_1$ and $\lambda_2$ are set to 0.2 and 1, respectively.

\subsection{\textbf{Render and Compare for Fast Scene Updating}}

In changeable environments where objects may be moved, resulting in unknown translations and rotations, we here propose a method that uses sparse view images, $\{I_{mov}\}$ (three for experiment), to predict the object's translation and rotation. These predictions are utilized as optimization parameters.
Specifically, we begin by applying MOG2 algorithm~\cite{zivkovic2004improved}\footnote{We use this algorithm, as it is simple and good enough in our task.} to detect moved pixels between the initial and current frames, whose center coordinates in current frames is denoted as $\{d_{gt}\}$. Next, we compute the mean semantic features of these pixels and use their cosine similarity to identify the 3D Gaussians of the moved objects, $\{g_i\}$. During optimization, we adjust $\{g_i\}$ using the predicted translation and rotation, while keeping other Gaussians fixed. Finally, we render two images: $I_{obj}$ with the adjusted Gaussians and $I_{pred}$ with all Gaussians, and define the optimization  $\mathcal{L}$ as,
\begin{align}
    \mathcal{L} &= \mathcal{L}_1(I_{mov}, I_{pred}) + \lambda_3\mathcal{L}_1(d_{pred}, d_{gt})
\end{align}
where $d_{pred}$ is the predicted location of the object on $I_{obj}$. We set the $\lambda_3$ to 0.1 to balance the loss of pixels and 2D distance. Since only the translation and rotation parameters of the objects require optimization, the process is efficient and typically completes in around 200 ms.

\subsection{\textbf{Language-guided Robotic Grasping}}
\label{sec:robot_grasping}
\begin{figure}[htbp]
\begin{center}
\includegraphics[width=0.9\linewidth]{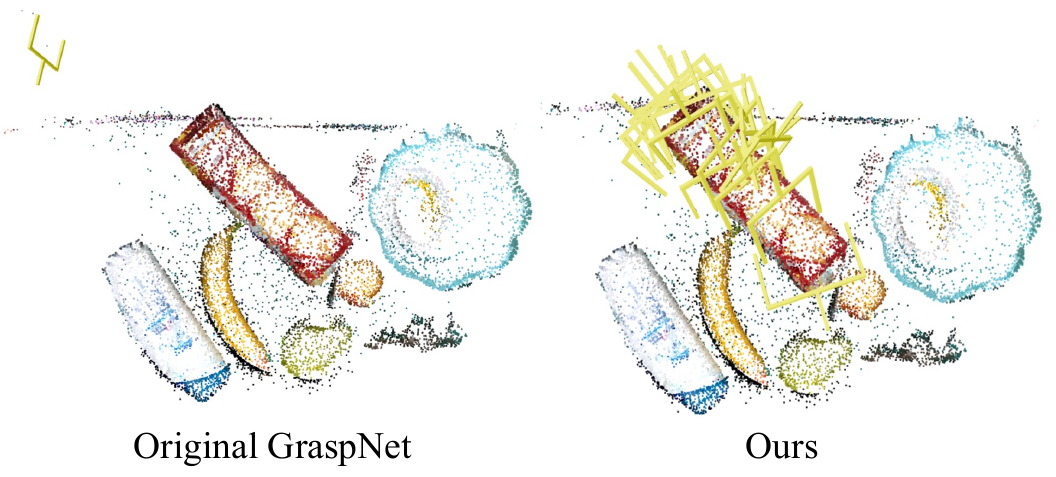}
\end{center}
\caption{Effectiveness of our grasp model: Unlike original GraspNet, which failed to generate grasp poses using 3DGS's centers, our model successfully generates grasp poses.}
\label{fig:grasp}
\vspace{-0.1in}
\end{figure}


In language-guided robotic grasping, existing methods~\cite{shen2023Distilled, lerftogo2023} often require processes like voxelization or multi-view depth and RGB information for scene reconstruction, which can introduce approximation errors. Unlike these methods, which resample the scene, our approach avoids direct use of sparse 3D Gaussians as point clouds for GraspNet, as illustrated in Fig.~\ref{fig:grasp}. Instead, we retrain GraspNet using $p_i, s_i, q_i$ as inputs. For training, we modify the original GraspNet dataset by selecting 100 scenes, segmenting objects and backgrounds, and reconstructing the scenes separately using RGB images of either objects or backgrounds. We then combine these to label the 3D Gaussians from the object segments as `objectness'. To account for real-world noise, we randomly add some 3D gaussian noise to each scene and use varying densification thresholds to enhance robustness.


\begin{figure*}[htbp]
\begin{center}
\includegraphics[width=0.8\linewidth]{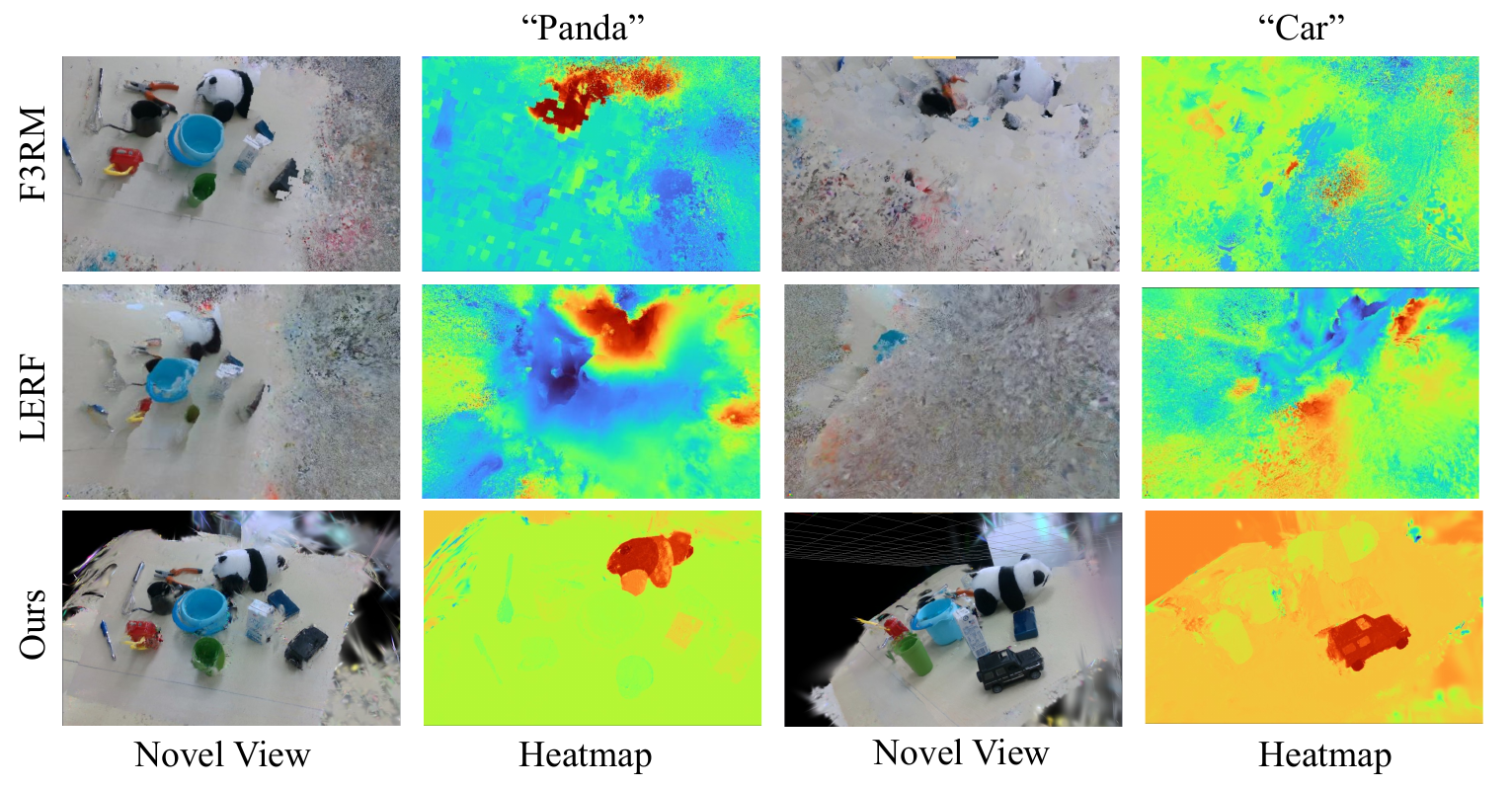}
\end{center}
\vspace{-0.15in}
\caption{Qualitative results of reconstruction and semantic distillation results.}\label{fig:sparse_compare}
\end{figure*}

\section{Experiments}


\noindent \textbf{Environment and Setup.}
In our physical robotic experiment, we use a KINOVA Gen2 robot with a 6-DoF curved wrist and a KG-3 gripper. The robot is equipped with three common cameras with good RGB images. 
The system runs on a desktop with an NVIDIA GTX A6000 GPU.

\noindent\textbf{Implementation Details.} To reconstruct the scene at a real-world scale, we use the calibrated camera poses as input to DUSt3R. The image resolution is also resized to 336 to serve as input for the MaskCLIP ViT-B/16 model, which extracts language features from each image. For 2D mask segmentation, we employ the SAM ViT-H model. Besides, we resample the initial point cloud to approximately 100,000 points and jointly train our 3DGS model while distilling semantic features. The pipeline is trained over 7,000 iterations, taking about 4 minutes. Due to the good initialization, the densification interval is set to 200 iterations.

\begin{figure*}[htbp]
\begin{center}
\includegraphics[width=0.85\linewidth]{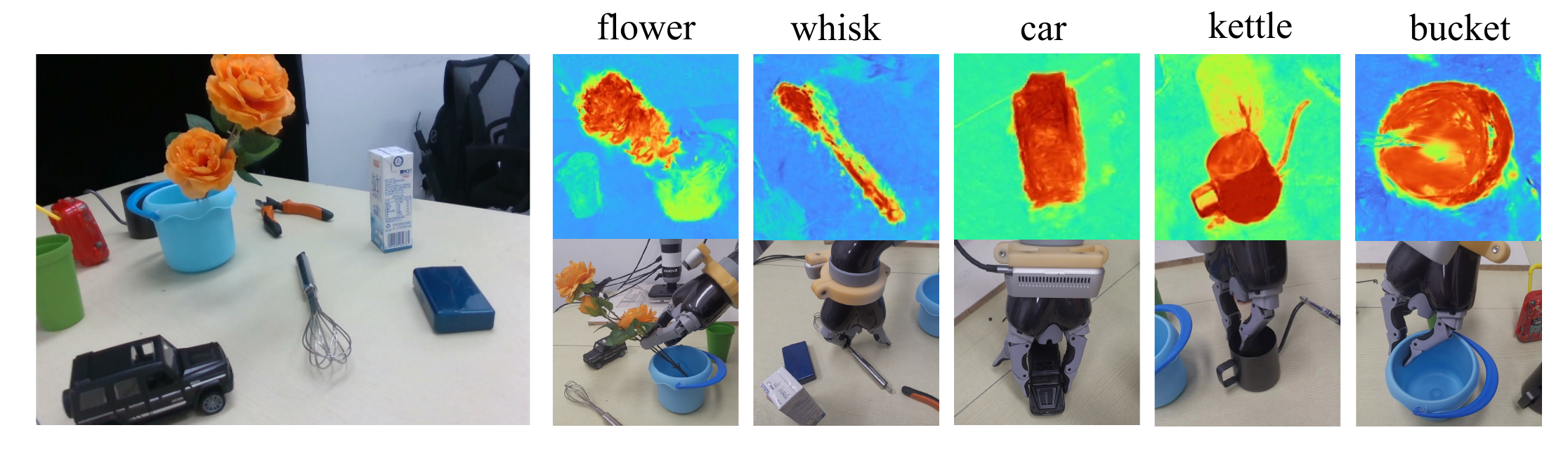}
\end{center}
\vspace{-0.1in}
\caption{Qualtitative results of language-guided grasping in the static environment. (Top Row) Heatmaps of given text queries. (Bottom Row)  Robot executing grasps sequentially without rescanning. }\label{fig:qual_grasp}
\label{fig:grasp_result}
\end{figure*}

\subsection{\textbf{Results of Reconstruction and Semantic Distillation}}

We compare our method with other state-of-art methods such as F3RM~\cite{shen2023Distilled} and LERF-TOGO~\cite{lerftogo2023} in terms of scene reconstruction and semantic distillation. Our method, F3RM and LERF are trained with 3 RGB images. Under this condition, both F3RM and LERF-TOGO exhibit overfitting. As illustrated in Fig.~\ref{fig:sparse_compare}, these methods fail to maintain the geometric integrity of objects, let alone produce reliable semantic outcomes.


\begin{table}
\small 
\centering
\caption{Quantitative results of success rate by using estimated grasping poses}\label{tab:static_grasp} 
\resizebox{1.0\linewidth}{!}{
\begin{tabular}{lcccccccc}
\toprule
& Flower & Whisk & Dragon & Car & Kettle's Lip & Panda & ScrewDriver & Total \\
\midrule
F3RM(3 Views) & 1/10 & 0/10 & 1/10 & 1/10 & 0/10 & 0/10 & 1/10 & 4/70 \\
LERF-TOGO(3 Views) & 0/10 & 1/10 & 0/10 & 1/10 & 1/10 & 1/10 & 1/10 & 5/70 \\
\midrule
F3RM(17 Views) & 6/10 & 1/10 & 5/10 & 7/10 & 7/10 & 5/10 & 7/10 & 38/70 \\
LERF-TOGO(17 Views) & 6/10 & 3/10 & 4/10 & 5/10 & 5/10 & 3/10 & 7/10 & 33/70 \\
\midrule
Ours(3 Views) & \textbf{8/10} & \textbf{8/10} & \textbf{9/10} & \textbf{8/10} & 7/10 & \textbf{8/10} & 7/10 & \textbf{55/70} \\
\bottomrule
\end{tabular}}
\end{table}

\begin{figure}
\begin{center}
\includegraphics[width=\linewidth]{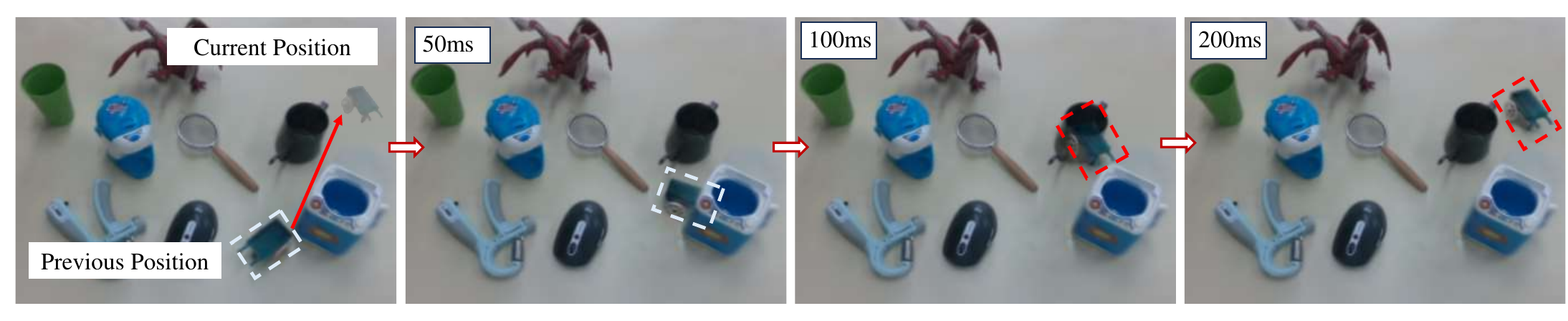}
\end{center}
\vspace{-0.1in}
\caption{Optimization process of our Render\&Compare module. We can generate the accurate position of the cart fastly. }\label{fig:dynamic}
\end{figure}

 



\subsection{\textbf{Results of Language-guided Grasping.}}

We first present results for language-guided grasping in static environments, followed by validation of our render-and-compare method for scene updating.
\begin{table}[htbp] \small  
\centering
\caption{\centering Result of grasping accuracy after movements. \label{tab:moving_object} }
\resizebox{1.0\linewidth}{!}{
 \begin{tabular}{ c  c  c  c  c  c  c } 
 \toprule
Methods & Time  & Scene1 & Scene2 & Scene3 & Scene4 & Scene5 \\  \midrule

F3RM~\cite{shen2023Distilled} & 10min & 4/5 & \textbf{4/5} & \textbf{5/5} & \textbf{4/5} & 3/5   \\
LERF-TOGO~\cite{lerftogo2023} & 34min & 3/5 & 4/5 & 4/5 & 3/5 & 3/5  \\
\midrule
\multirow{3}{*}{Ours} & 50 ms   & 0/5 & 0/5 & 0/5 & 1/5 & 0/5 \\
     & 100 ms & 3/5 & 2/5 & 3/5 & 2/5 & 3/5  \\
     & 200 ms & \textbf{5/5} & \textbf{4/5} & \textbf{5/5} & \textbf{4/5} & \textbf{4/5} \\
 \bottomrule
\end{tabular}}
\vspace{-0.1in}

\end{table}

\noindent{\textbf{Results of Grasping Accuracy in Static Environment.}} 
As shown in Tab.~\ref{tab:static_grasp}, both F3RM and LERF-TOGO demonstrate poor grasping performance in sparse views, and even in dense views, they struggle to successfully grasp the whisk. Their failure in sparse views arises from their inability to accurately reconstruct the scene. Additionally, the difficulty in grasping the whisk can be attributed to the implicit representation used by NeRF, which requires voxelization to generate the 6D pose. This voxelization process brings a loss of precision. In contrast, our method achieves high-precision grasping across all objects, even in the sparse view setting. This is due to its superior reconstruction accuracy and more effective semantic distillation. Additionally, our language-guided grasping results are presented in Fig.~\ref{fig:grasp_result}.



\noindent{\textbf{Results of Grasping Accuracy in Changeable Scene.}}
In this scenario, objects could be randomly shifted by people. Rather than necessitating dense view images for scene updates like F3RM and LERF-TOGO, our approach employs Render-and-Compare for quicker updates. As illustrate in Tab.~\ref{tab:moving_object}, F3RM and LERF-TOGO are unable to update specific areas of the scene, requiring a full scene reconstruction instead. Note that the times listed for F3RM and LERF-TOGO in Tab.\ref{tab:moving_object} exclude the additional 2 minutes required for image capturing, a step our method does not require. In contrast, our render-and-compare strategy can quickly update the scene, taking approximately 200ms.



\noindent{\textbf{Result of Semantic Distillation Effect.}}
We use 2D IoU as a metric to evaluate the effectiveness of our semantic distillation in both training and novel views. Additionally, we compare our results with other methods using either sparse or dense views, as shown in Tab.~\ref{tab: 2D IOU}. F3RM and LERF-TOGO fail to preserve object geometries, leading to significantly lower IoU scores in novel views. In contrast, our approach excels at maintaining geometric integrity across novel views, demonstrating its ability to infuse semantic information into 3D objects.


\begin{table}[hbtp] \small
\centering
\caption{\centering Quantitative results of 2D IOU. \label{tab: 2D IOU} }
\vspace{-0.1in}
\resizebox{\linewidth}{!}{
 \begin{tabular}{ c  c  c  c  c  c  c } 
 \toprule
 
Methods & \makecell[c]{F3RM\\(3 view)} & \makecell[c]{LERF-TOGO\\(3 view)} & \makecell[c]{F3RM\\(17 view)} & \makecell[c]{LERF-TOGO\\(17 view)} & \makecell[c]{Ours\\(3 view)} \\  \midrule

Training Views & 0.75 & 0.69 & 0.81 & 0.74 & \textbf{0.83} \\
Novel Views   & 0.13 & 0.08 & \textbf{0.75} & 0.71 & 0.71\\
 \bottomrule
\end{tabular}}
\vspace{-0.1in}
\end{table}

\noindent{\textbf{Results of Time Consumption.}} 
In Tab.~\ref{tab:time}, show the results of all methods in both static scenes and changeable environments. In contrast, our SparseGrasp not only significantly outperforms F3RM and LERF-TOGO, but also is much faster than LangSplat, which requires extensive preprocessing time. Furthermore, when the scene changes, these methods need to recapture images and reconstruct the entire scene, which is time-consuming. While our can quickly update the changed objects, offering a more efficient solution. 



\begin{table}[hbtp] \small 
\caption{\centering Comparison of Time Consumption.\label{tab:time}}
\centering
\resizebox{1.0\linewidth}{!}{
\begin{tabular}{ccccc}
\toprule
Methods & F3RM & LERF-TOGO & LangSplat & Ours\\ 
\midrule
Static Scene & 10min & 34min & 3600min+ & \textbf{4min} \\
Scene Updation & - & - & - & 200ms \\
\bottomrule 
\end{tabular}
}
\vspace{-0.15in}
\end{table}

\subsection{More Analysis}
\begin{figure} 
\begin{center}
\includegraphics[width=1.0\linewidth]{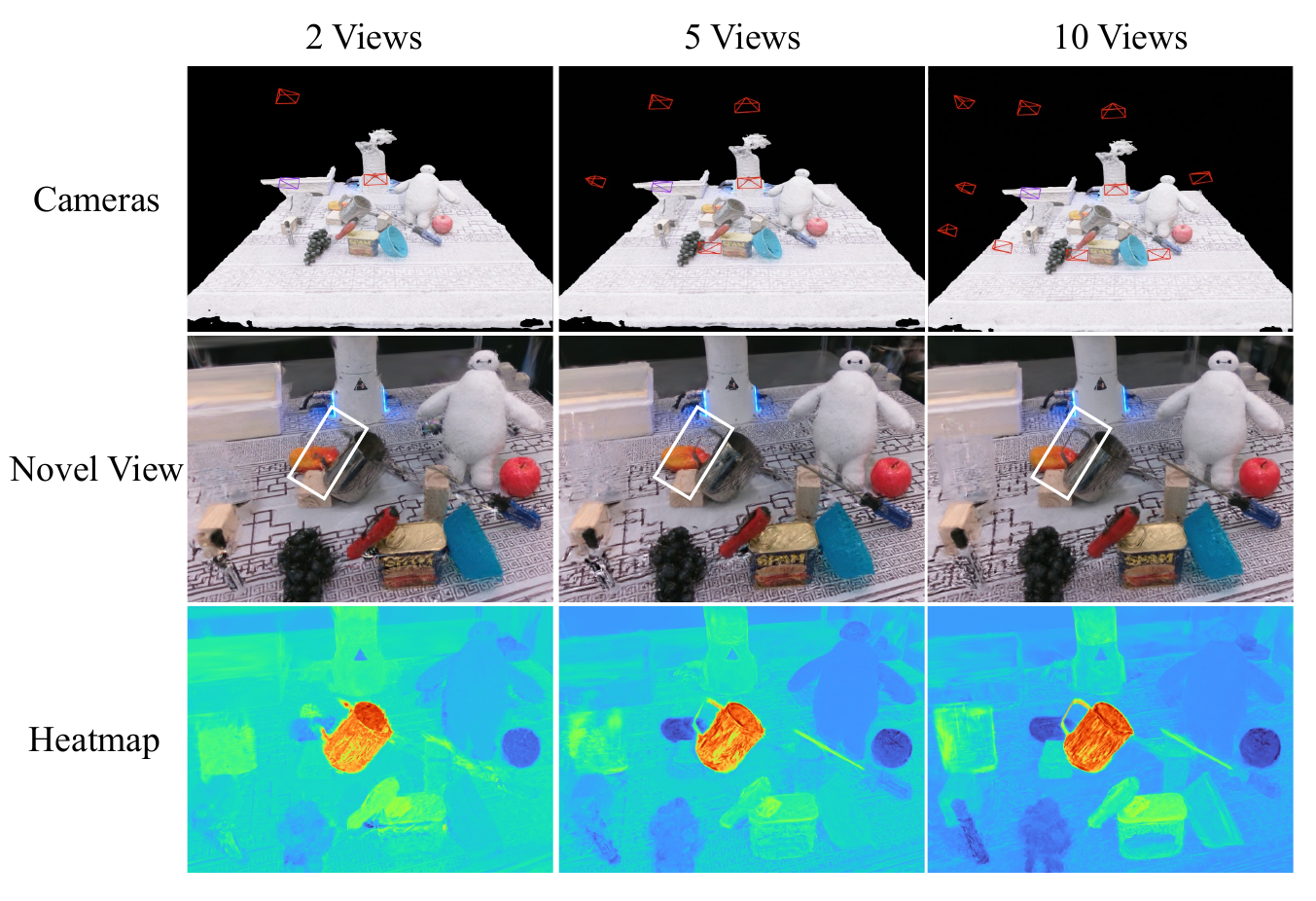}
\end{center}
\vspace{-0.1in}
\caption{Rendered images and heatmaps for the query `metal hug' across different camera views. Red and purple indicate training and testing views, respectively (top row).
 }\label{fig:more_less_view}
\end{figure}

\noindent{\textbf{Results on the Different Number of Views.}} 
Fig.~\ref{fig:more_less_view} shows a qualitative comparison of RGB images and heatmaps of the `metal mug' from the novel view in F3RM dataset. As the number of training images increases, the boundary around the slim part of the metal mug becomes noticeably clearer. Additionally, Even with only two views, our method is still able to accurately distinguish the object.


\begin{figure}[htbp]
\vspace{-0.1in}
\begin{center}
\includegraphics[width=\linewidth]{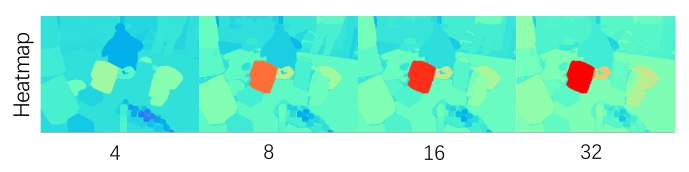}
\end{center}
\caption{Qualitative results of heatmaps when 
applying PCA with different numbers of components .}
\label{fig:pca}
\vspace{-0.2in}
\end{figure}

\noindent{\textbf{Results of Different Number of PCA's Components.}} 
To evaluate the compression ability of PCA, we explore varying numbers of PCA components. As shown in Fig.~\ref{fig:pca}, we visualize the heatmaps of the metal mug with different numbers of PCA components. We observe that using as few as 16 components already achieves effective segmentation of the metal mug.

\noindent \textbf{Discussion and Future Work.}
\textit{1) Can SparseGrasp handle adding or replacing objects in the scene?} Yes, but requires a `refresh'. Our system primarily focuses on multi-turn grasping in changeable environments. While it needs re-initialization when adding or removing objects, it still outperforms \textit{F3RM} and \textit{LERF-TOGO} in speed, as shown in Table~\ref{tab:time}. In future work, we aim to explore 3D inpainting methods~\cite{ye2023gaussian} to improve processing speed for newly added or removed objects.
\textit{2) How about directly using optical flow or tracking?} Possible, but depth estimation and semantic integration are still needed. Some works~\cite{bharadhwaj2024track2act, li2024objectaware} use tracking or optical flow for scene updation, but these methods rely on depth data, which is difficult to estimate from sparse RGB views and increases processing time. Moreover, they lack the semantic understanding necessary for text-based queries like ours. Our method enables fast scene updates using only sparse RGB images, without requiring depth estimation, while maintaining semantic capabilities.

\section{Conclusions}

We propose \textbf{SparseGrasp}, a system for rapid scene reconstruction, understanding and updating using sparse-view RGB images. By integrating \textit{DUSt3R} for dense point cloud initialization and \textit{MaskCLIP} with \textit{SAM} for efficient semantic extraction, we overcome the limitations of dense-view dependencies. Dimensionality reduction via \textit{PCA} boosts efficiency, while our render-and-compare strategy enables fast scene updates without full reconstructions. Finally, our 3DGS-based grasping method streamlines grasp generation, avoiding voxelization issues. \textbf{SparseGrasp} improves scene understanding, speed and robustness, paving the way for future advancements.




\bibliographystyle{IEEEtran}
\bibliography{root}

\end{document}


\maketitle








\section{More Analysis}

\noindent\textbf{Reconstruction and semantic distillation Time.} \\
Compared to those who use depth sensor, current RGB-only methods can be time-consuming. Our method greatly reduce the time consumption, as can be shown in Tab \ref{tab: time compare}. Our reconstruction and semantic distillation time can be even shorter than original 3D GS who only reconstruct RGB information. Furthermore, our methods can swiftly adjust to scene changes in \textit{near real-time.}

\begin{table}[hbtp] \small
\centering
\caption{\centering Result of reconstruction and semantic distillation time and quality. \label{tab: time compare} }

 \begin{tabular}{ c  c  c  c  c  c  c } 
 \toprule
 
Methods & F3RM~\citep{shen2023Distilled} & LERF-TOGO~\citep{lerftogo2023} & LangSplat~\citep{qin2023langsplat} & Original 3DGS(Only RGB) & Ours \\  \midrule

reconstruction time & 622s & 1932s & 5h+ & 450s & \textbf{180s} \\
 \bottomrule
\end{tabular}
\vspace{-0.15in}

\end{table}

\noindent\textbf{Result of semantic distillation effect.}

Here we use 2D IOU as metrics, testing our semantic distillation effects in learned views as well as novel views. We also compared our results with others using either sparse views or dense views. Results are shown in Tab\ref{tab: 2D IOU}.

\begin{table}[hbtp] \small
\centering
\caption{\centering 2D IOU Results. \label{tab: 2D IOU} }

 \begin{tabular}{ c  c  c  c  c  c  c } 
 \toprule
 
Methods & \makecell[c]{F3RM\\(3 view)} & \makecell[c]{LERF-TOGO\\(3 view)} & \makecell[c]{F3RM\\(17 view)} & \makecell[c]{LERF-TOGO\\(17 view)} & \makecell[c]{Ours\\(3 view)} \\  \midrule

Learned Views & 0.75 & 0.69 & 0.81 & 0.74 & \textbf{0.83} \\
Novel Views   & 0.13 & 0.08 & \textbf{0.75} & 0.71 & 0.71\\
 \bottomrule
\end{tabular}
\vspace{-0.15in}
\end{table}
As evident in Tab\ref{tab: 2D IOU}, F3RM and LERF-TOGO do not maintain object geometries, resulting in notably poor IOU outcomes for novel perspectives. In contrast, our approach excels in preserving such aspects under new viewpoints, suggesting its ability to imbue semantic information into 3D objects.

\noindent\textbf{Ablation Results.} \\
Here, We utilize different loss functions to forecast objects' movements randomly manipulated by humans. We keep using grasping accuracy as a vivid metrics for validating whether predicting is accurate. From the table\ref{tab: diff loss}, we can find that: in condition 1), without $\mathcal{L}_{obj}$, training is difficult to converge, and the success rate of grasping is very low. The reason for this is that the loss does not necessarily decrease as the object gets closer to its actual position. In fact, from the intermediate results, this is more like random drift. In condition 2), the grasping accuracy is higher than one. This is because with $\mathcal{L}_{obj}$, a rough estimation on movement can be ensured. However, without $\mathcal{L}_{fg}$, estimation is not precise and can lead to errors in extreme cases.

\begin{table}[hbtp] \small
\centering
\caption{\centering Result of Grasping accuracy Using different Loss.}

 \begin{tabular}{ c  c  c  c c} 
 \toprule
 Loss & Time(ms) & Dragon & Flower & Panda \\ \midrule         
   \multirow{2}{*}{w/o $\mathcal{L}_{obj}$} & 30  & 1/5 & 0/5 & 0/5 \\
         & 50  & 1/5 & 1/5 & 0/5 \\ \midrule
   \multirow{2}{*}{w/o $\mathcal{L}_{fg}$} & 30  & 0/5 & 1/5 & 1/5 \\
         & 50  & 2/5 & 1/5 & 1/5 \\ 

 \bottomrule
\end{tabular}
\vspace{-0.15in}
\label{tab: diff loss}
\end{table}

\noindent\textbf{Grasping transparent and metal objects.} \\
In the main text, we mentioned that compared to using a depth sensor, using pure RGB images for capturing transparent and metallic objects has more advantages. In the experimental section of the main text, to highlight the versatility of our method, we did not conduct many related experiments. Here we provide additional information to supplement this. We constructed scenes composed of metallic objects and transparent objects for the grasping experiment. One of the scene is shown as Fig.~\ref{fig:trans}. Other scenes combine metal and transparent objects with other kind of objects.

\begin{figure*}[htbp]
\begin{center}
\includegraphics[width=0.5\linewidth]{pic/trans.pdf}
\end{center}
\vspace{-0.1in}
\caption{One of the scene for testing transparent and metal objects.}\label{fig:trans}
\end{figure*}

\begin{figure*}[htbp]
\begin{center}
\includegraphics[width=\linewidth]{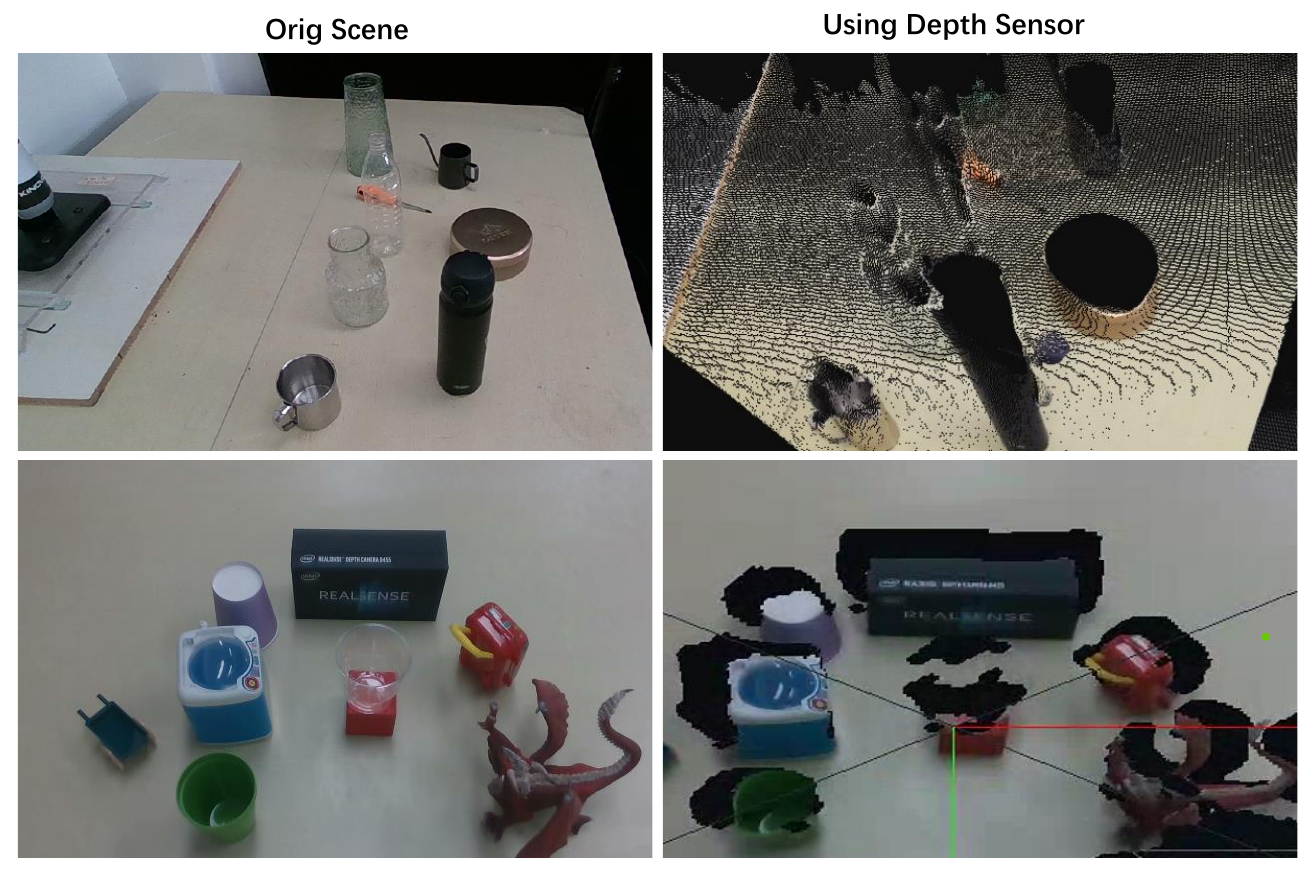}
\end{center}
\vspace{-0.1in}
\caption{One of the scene for testing transparent and metal objects.}\label{fig: sensor}
\end{figure*}

Our grasping results are displayed in Tab. \ref{tab: trans}. Demonstrably, our method adeptly handles these types of objects. Despite testing methods with a depth sensor, they consistently falter in capturing the scene's point cloud, as illustrated in Fig. \ref{fig: sensor}, let alone succeeding in subsequent grasping attempts.

\begin{table}[hbtp] \small
\centering
\caption{\centering Result of grasping transparent and metal objects. \label{tab: trans} }

 \begin{tabular}{c c c c c} 
 \toprule
 
Methods & Scene1 & Scene02 & Scene03 & Sum \\  \midrule

Grasping Accuracy & 7/10 & 6/7 & 6/8 & 19/25 \\
 \bottomrule
\end{tabular}
\vspace{-0.15in}

\end{table}

\bibliography{example}  